\documentclass[runningheads]{llncs}
\usepackage{graphicx}
\usepackage[utf8]{inputenc}
\usepackage{float}
\usepackage{amsmath,amssymb,amsfonts}
\usepackage{xcolor}
\usepackage{colortbl}
\usepackage{hhline}
\usepackage{booktabs}
\usepackage[numbers,sectionbib]{natbib}
\usepackage{tikz}
\usepackage{rotating}
\usepackage{graphbox}
\usepackage{csvsimple}

\newcommand{\plus}{\oplus}
\newcommand{\minus}{\ominus}
 
\newcommand{\given}{\ |\ }

\newcommand{\nrtrees}{K}

\newcommand{\leafs}{\mathcal{W}}
\newcommand{\leaf}{\mathbf{w}}
\newcommand{\lpos}{w^{\plus}}
\newcommand{\lneg}{w^{\minus}}

\newcommand{\predfkt}{p}
\newcommand{\pprob}{\predfkt_{prob}}

\newcommand{\plap}{\predfkt_{lap}}

\newcommand{\pcb}{\predfkt_{cb}}
\newcommand{\ppls}{\predfkt_{pls}}

\newcommand{\pvec}{\mathbf{v}}
\newcommand{\pelem}{v}

\newcommand{\aggfkt}{h}
\newcommand{\aavg}{\aggfkt_{avg}}

\newcommand{\avote}{\aggfkt_{vote}}

\newcommand{\predaggfkt}{g}
\newcommand{\fvec}{\felem_1, \felem_2, \ldots, \felem_\nrtrees}
\newcommand{\felem}{\leaf}
\newcommand{\fpool}{\predaggfkt_{pool}}

\newcommand{\fdem}{\predaggfkt_{dempster}}
\newcommand{\fcau}{\predaggfkt_{cautios}}

\newcommand{\feva}{\predaggfkt_{eva}}

\newcommand*\circled[1]{\tikz[baseline=(char.base)]{
            \node[shape=circle,draw,inner sep=1pt] (char) {#1};}}

\usepackage[misc]{ifsym}
\usepackage{xcolor}
\newcommand{\com}[1]{{\leavevmode\color{blue}COM: #1}}
 \renewcommand{\com}[1]{}
\newcommand{\todo}[1]{{\leavevmode\color{red}TODO: #1}}
 \renewcommand{\todo}[1]{}
\newcommand{\remove}[1]{}

\begin{document}
\title{Combining Predictions under Uncertainty: \\The Case of Random Decision Trees}
\titlerunning{Combining Predictions under Uncertainty: Random Decision Trees}

\author{Florian 
Busch\inst{1}
\and Moritz Kulessa\inst{1} \and
 Eneldo Loza Menc\'{i}a\inst{1} 
\and
Hendrik Blockeel\inst{2}
}

\institute{Technische Universit\"at Darmstadt, Darmstadt, Germany \\
\email{\{florianpeter.busch@stud, mkulessa@ke, eneldo@ke\}.tu-darmstadt.de}
\and
KU Leuven, Leuven, Belgium \\
\email{hendrik.blockeel@kuleuven.be}}
\maketitle              
\begin{abstract}

A common approach to aggregate classification estimates in an ensemble of decision trees is to either use 
voting or to average the probabilities for each class.
The latter takes uncertainty into account, but not the reliability of the uncertainty estimates (so to say, the ``uncertainty about the uncertainty’’).
More generally, much remains unknown about how to best combine probabilistic estimates from multiple sources.
In this paper, we investigate a number of alternative prediction methods. Our methods are inspired by the theories of  probability, belief functions and reliable classification, as well as a principle that we call \emph{evidence accumulation}. 
Our experiments on a variety of data sets are based on random decision trees which guarantees a high diversity in the predictions to be combined.
Somewhat unexpectedly, we found that taking the average over the probabilities is actually hard to beat. 
However, evidence accumulation showed consistently better results on all but very small leafs.

\end{abstract}

\section{Introduction \com{until page 2. Status: almost final}}
\label{sec:introduction}


Ensemble techniques, such as bagging or boosting, are popular tools to improve the predictive performance of classification algorithms due to 
diversification and stabilization of predictions~\citep{ensemble_dietterich}. Beside learning the ensemble, a particular challenge is to combine the estimates of the single learners during prediction. The conventional approaches for aggregation are to follow the majority vote or to compute the average probability for each class. More sophisticated approaches are to use \emph{classifier fusion} \citep{fusion_classifier}, \emph{stacking} \citep{stacking} or \emph{mixture of experts}~\citep{yuksel2012twenty}. Most of these approaches mainly aim to identify strong learners in the ensemble, to give them a higher weight during aggregation. However, this ignores the possibility that the predictive performance of each learner in the ensemble may vary depending on the instance to be classified at hand. 
To take these differences into account, we argue that considering the certainty and uncertainty of each individual prediction might be a better strategy to appropriately combine the estimates.
This should be particularly the case for ensembles generated using randomization techniques since this potentially leads to many unstable estimates. 

A popular and effective strategy is to randomize decision trees. The prediction of a single decision tree is usually based on statistics associated with the leaf node to which a test instance is forwarded. 
A straight-forward way of assessing the reliability of a leaf node's estimate is to consider the number of instances which have been assigned to the node during learning.
For example, imagine a leaf containing 
$4$ instances of class \emph{A} and 0 of \emph{B}, and another leaf with $10$ instances in class \emph{A} and $40$ in class \emph{B}. 
The first source of prediction has considerably less  evidence to \emph{support}
its prediction in contrast to the second leaf. 
If both leafs need to be combined during prediction, it seems natural to put more  trust on the second leaf and, therefore, one may argue that class \emph{B} should be preferred though the average probability of $0.6$ for \emph{A} tells us the opposite. 
However, one may argue for \emph{A} also for another reason, namely the very high confidence in the prediction for \emph{A}, or put differently, the absence of uncertainty in the rejection of \emph{B}.
The certainty of the prediction is also known as \emph{commitment} in the theory of belief functions.
In this work, we will review and propose techniques which are grounded in different theories on modelling uncertainty and hence take the 
\emph{support and strength of evidence}
 differently into account.

More specifically, \citet{shaker2020aleatoric} introduced a formal framework which translates observed counts as encountered in decision trees into a vector of  plausibility scores for the binary classes, which we use for the prediction, and scores for  aleatoric and epistemic uncertainty.
While 
aleatoric uncertainty captures the
randomness inherent in the sample, the latter captures the uncertainty with regard to the lack of evidence. 
The theory of belief functions \citep{shafer1976mathematical}  contributes two well-founded approaches to our study in order to combine these uncertainties.
In contrast to the theory of belief functions, in probability theory the uncertainty is usually expressed by the dispersion of the fitted distribution. 
We introduce a technique which takes advantage of the change of the shape of the distribution in dependence of the sample size.
And finally, apart from combining the probabilities themselves, one can argue that {\em evidence} for a particular class label should be 
properly \emph{accumulated}, rather than averaged across the many members of the ensemble.
In this  technique, we measure the strength of the evidence by how much the probabilities deviate from the prior probability.


As an appropriate test bed for combining uncertain predictions, we choose to analyse these methods based on random decision trees~(RDT)~\citep{rdt_fan}. 
In contrast to other decision tree ensemble learners, such as random forest, RDT do not optimize any objective function during learning, which results in a large diversity of class distributions in the leafs.
This puts the proposed methods specifically to the test in view of the goal of the investigation and makes the experimental study independent of further decisions such as the proper selection of the splitting criterion.
We have compared the proposed techniques on $21$ standard binary classification data sets. 
Surprisingly, our experimental evaluation showed that 
methods, which take the amount of evidence into consideration, did only improve over the simple averaging baseline in some specific cases. However, we could observe an advantage for the proposed approach of \emph{evidence accumulation} which takes the strength of the evidences into account.

\section{Preliminaries \com{1-2 pages. Status: almost final}}
\label{sec:preliminaries}


This section briefly introduces RDT, 
followed by a short discussion on previous work which is relevant to us. Throughout the paper, we put our focus on binary classification which is the task of learning a mapping 
$f: \mathrm{X}^m \rightarrow y$ between an instance $x \in \mathrm{X}^m$ with  $m$ numerical and/or nominal features and a binary class label $y \in \{\minus,\plus\}$ through a finite set of observations $\{(x_1,y_1), \ldots, (x_n,y_n)\}$.


\subsection{Random Decision Trees}

Introduced by~\citet{rdt_fan}, the approach of RDT is an ensemble of randomly created decision trees which, in contrast to classical decision tree learners and random forest~\citep{random_forest}, do not optimize a objective function during training. More precisely, the inner tests in the trees are chosen randomly which reduces the computational complexity but still achieves competitive and robust performance.


\subsubsection{Construction.}
Starting from the root node, inner nodes of a single random tree are constructed recursively by distributing the training instances according to the randomly chosen test at the inner node 
as long as the stopping criterion of a minimum number of instances for a leaf is not fulfilled. Discrete features are chosen without replacement for the tests in contrast to continuous features, for which additionally a randomly picked instance determines the threshold. 
In case that no further tests can be created, a leaf will be constructed in which information about the assigned instances will be collected. For binary classification, the number of positive $\lpos$ and negative $\lneg$ class labels are extracted. Hence, a particular leaf can be denoted as $\leaf = [\lpos, \lneg]$ where $\leaf\in \leafs$ and $\leafs = \mathbb{N}^2$.

\subsubsection{Prediction.} 
For each of the $\nrtrees$ random trees in the ensemble, the instance to be classified is forwarded from the root to a leaf node passing the respective tests in the inner nodes. The standard approach  mainly used in the literature to aggregate the assigned leaf nodes $\leaf_{1}, \dots, \leaf_{\nrtrees}$ is to first compute the probability for the positive class on each leaf which is then averaged across the ensemble. 



\subsection{Related Work \com{0.5-0.75 pages. Status: some work-over and maybe more meaningful work} }


A very well known approach used for improving the probability estimates of decision trees is Laplace smoothing,  
which essentially incorporates epistemic uncertainty through a prior. 
However, whereas \citet{provost2003tree} showed a clear advantage over using the raw estimates for single decision trees, \citet{prob_random_forest} observed that for random forest better estimates are achieved without smoothing.
Apart from that correction, the reliability of the individual predictions in decision trees is usually controlled via pruning or imposing  leaf sizes \citep{provost2003tree,zhou2021trees}.  
While the common recommendation is to grow trees in an ensemble to their fullest extent, \citet{zhou2021trees} argue that depending on the properties of the underlying data greater leafs and hence more stable predictions are preferable. 
\com{if necessary, include here short discussion on well calibrated PETs}
With respect to RDT, \citet{kulessa2018dynamic} reward predictions with higher confidence in the ensemble using the inverted Gini-index. However, their weighting approach has not been directly evaluated.

The Dempster-Shafer theory of evidence, or theory of belief functions, \citep{shafer1976mathematical} is also concerned with the strength of the confidences. 
The general framework for combining prediction of \citet{lu1996knowledge} is based on this theory and the study showed that the proposed technique can improve upon the Bayesian approach. 
\citet{raza2006classifier} similarly combined outputs of support vector machines but only compared to a single classifier in their evaluation. 
\citet{nguyen2018aggregation} take uncertainties in an alternative way into account by 
computing interval-based information granules for each classifier. 
They could improve over common ensemble techniques, however, their approach requires 
additional optimization during learning and during classification.
As already discussed in Section~\ref{sec:introduction}, meta and fusion approaches also often require additional learning steps and ignore the uncertainty of individual predictions.
\citet{costa2018combining} also ignore the uncertainties but propose interesting alternative aggregations for ensembles based on generalized mixture functions.\com{perhaps add some information on our work on this.}

\section{
Aggregation of Scores from Leafs
\com{2 pages. Status: only little work-over}
}
\label{sec:gradual}

Based on the leaf nodes $\leaf_{1}, \dots, \leaf_{K}$ to which an instance has been assigned to during prediction, the concept explained in this section is to first convert the leafs into scores which are then combined using aggregation functions.



\subsection{Scoring Methods}
\label{sec:dt_pred}

\begin{figure}[b!]
\includegraphics[width=\textwidth]{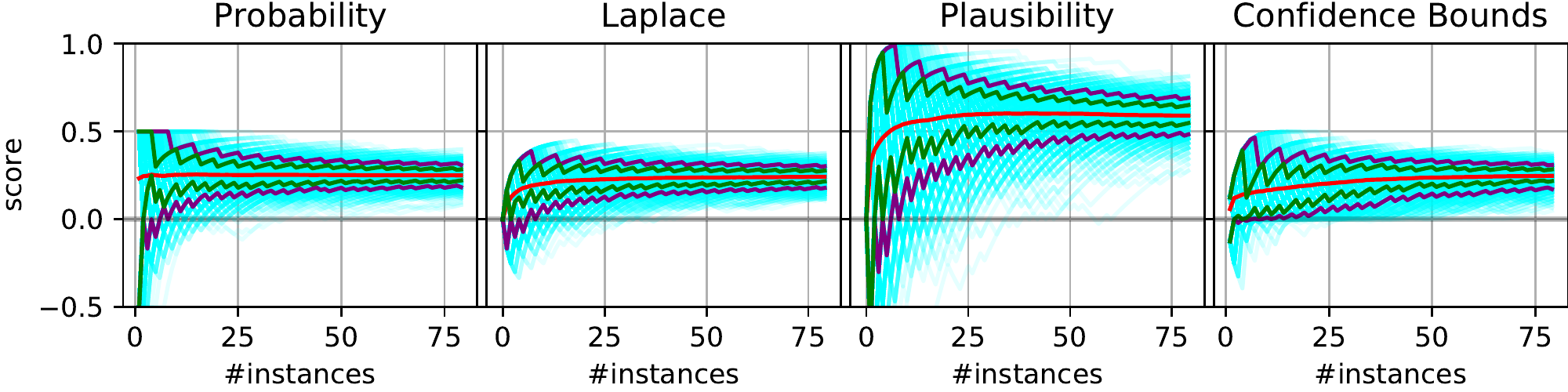} 
\caption{Evolution of the scores ($y$-axis) for each scoring method based on simulating a leaf to which random samples are added ($x$-axis) with a probability of $75\%$ for the positive class. This Bernoulli trial is repeated $100$ times (cyan lines). The average over the trials scores is depicted by the red line and the $10\%$, $25\%$, $75\%$ and $90\%$ quantiles by the green and purple line, respectively.}
\label{fig:prediction_methods}
\end{figure}

In this section, the scoring methods are introduced of which most are designed to take the uncertainty of the leafs into account. We denote $\predfkt: \leafs \rightarrow \mathbb{R}$ as the function which assigns a score $\pelem=\predfkt(\leaf)$ to a leaf $\leaf \in \leafs$ where $\leaf = (\lpos, \lneg)$. All scoring methods proposed in this work are designed such that the sign of the resulting score indicates whether the positive or the negative class would be predicted.
In order to understand how the scoring methods deal with uncertainty, we introduce an approach to visualize and compare them in Figure~\ref{fig:prediction_methods}.



\subsubsection{Probability and Laplace.}

Computing the probability $\pprob(\leaf)$ is the conventional approach to obtain a score from a leaf. Hence, it will serve as a reference point in this work.
In addition, we include the  \emph{Laplace} smoothing  $\plap(\leaf)$ which corrects the estimate for uncertainty due to a lack of samples.
\begin{align*}
    \pprob(\leaf) = \frac{\lpos}{\lpos+\lneg} - 0.5  && \plap(\leaf) = \frac{\lpos + 1}{\lpos+\lneg+2} - 0.5 
\end{align*}
In Figure~\ref{fig:prediction_methods}, we can see how both $\pprob$ and $\plap$ converge to the same score for larger leafs but that $\plap$ has much less extreme predictions for small leafs.



\subsubsection{Plausibility.}
\label{sec:plausibility}


The idea of this approach is to measure the uncertainty in terms of aleatoric and epistemic uncertainty. In order to define the uncertainty, \citet{shaker2020aleatoric} introduces the degree of support 
for the positive class $\pi(\plus|\leaf)$ and the negative class $\pi(\minus|\leaf)$ which can be calculated for a leaf $\leaf$ as follows:
\begin{equation*}
\begin{aligned}
\pi(\plus|\leaf) &= \sup_{\theta\in [0,1]} \min\left( \left(\frac{\theta}{(\frac{\lpos}{\lpos+\lneg})}\right)^{\lpos} \left(\frac{1-\theta}{(\frac{\lneg}{\lpos+\lneg})}\right)^{\lneg} ,2\theta-1\right)\text{,}\\
\pi(\minus|\leaf) &= \sup_{\theta\in [0,1]} \min\left( \left(\frac{\theta}{(\frac{\lpos}{\lpos+\lneg})}\right)^{\lpos} \left(\frac{1-\theta}{(\frac{\lneg}{\lpos+\lneg})}\right)^{\lneg} ,1-2\theta\right)\text{.}
\end{aligned}
\end{equation*}
Based on this \emph{plausibility} the epistemic uncertainty $u_e(\leaf)$ and the aleatoric uncertainty $u_a(\leaf)$ can be defined as:
\begin{align*}
    u_e(\leaf) = \min\left[\pi(\plus|\leaf),\pi(\minus|\leaf)\right] &&  u_a(\leaf) = 1-\max\left[\pi(\plus|\leaf),\pi(\minus|\leaf)\right]
\end{align*}
Following \citet{nguyen2018reliable}, the degree of preference for the positive class $s_\plus(\leaf)$ can be calculated as
\begin{equation*}
s_\plus(\leaf) = 
\begin{cases}
1-(u_a(\leaf) + u_e(\leaf))& \text{if } \pi(\plus|\leaf) > \pi(\minus|\leaf)\text{,}\\
\frac{1-(u_a(\leaf)+u_e(\leaf))}{2}& \text{if } \pi(\plus|\leaf) = \pi(\minus|\leaf)\text{,}\\
0& \text{if } \pi(\plus|\leaf) < \pi(\minus|\leaf)
\end{cases}
\end{equation*}
and analogously for the negative class.
We use the trade-off between the degrees of preference as our score:
\begin{equation*}
\ppls(\leaf) = s_\plus(\leaf) - s_\minus(\leaf)\text{.}
\end{equation*}

With respect to Figure~\ref{fig:prediction_methods}, we can observe that the plausibility approach models the uncertainty similarly to the Laplace method but also preserves a high variability for small leaf sizes. From that point of view, it could be seen as a compromise between the Probability and the Laplace method. 


\subsubsection{Confidence Bounds.}
\label{subsec:tree_cb}

Another possibility to consider uncertainty is to model the probability of the positive and the negative class each with a separate probability distribution and to compare these. 
In this work, we use a beta-binomial distribution parameterized with $\lpos+\lneg$ number of tries, $\alpha = \lpos+1$, and $\beta = \lneg+1$ to model the probability of the positive class, analogously for the negative class. 
To get a measure $c(\leaf)$ of how well both classes are separated, we take the intersection point in the middle of both distributions, normalized by the maximum height of the distribution, and use this as in the following: 
\begin{align*}
\pcb(\leaf) = (1-c(\leaf)) \left(\frac{\lpos}{\lpos+\lneg}-0.5\right)
\label{eq:dt_ucb}
\end{align*}
Hence, $c(\leaf)$ generally decreases with 1) increasing class ratio in the leaf, or 2) with more examples, since this leads to more peaky distributions.
In contrast to Laplace and Plausibility in Figure~\ref{fig:prediction_methods}, the scores of confidence bounds are increasing 
more steadily which indicates a weaker consideration of the size of the leafs than the aforementioned methods.

\subsection{Aggregation Functions}
\label{sec:aggregation}


In a final step the scores $\pelem_1, \ldots, \pelem_\nrtrees$, where $\pelem_i=\predfkt(\leaf_i)$, need to be combined using an aggregation function $\aggfkt: \mathbb{R}^\nrtrees \rightarrow \mathbb{R}$.
Due to space restrictions, we limit our analysis in this work to the arithmetic mean and 0-1-voting
\begin{align*}
    \aavg(\pvec) = \frac{1}{\nrtrees}\sum_{i=1}^{\nrtrees} \pelem_i && \avote(\pvec)=\frac{1}{\nrtrees} \sum_{i=1}^\nrtrees sgn(\pelem_i > 0)
\end{align*}
with $sgn(\cdot)$ as the sign function. 
We consider $\avote$ as a  separate method in the following, since all presented scoring methods change their sign equally.
Note that $\pprob$ in combination with $\aavg$ corresponds to what is known as \emph{weighted voting}, and $\pcb$ an instantiation of \emph{weighted averaging} over $\pprob$.

Similarly to \citet{costa2018combining}, we found in preliminary experiments that alternative approaches, including the median, maximum and a variety of generalized mixture functions \citep{farias2016some}, could be beneficial in some cases, but did not provide meaningful new insights in combination with the explored scoring functions.

\section{
Integrated Combination of Leafs
\com{3 pages. Status: some iterations}
}
\label{sec:other_methods}

\begin{figure}[b!]
\includegraphics[width=\textwidth]{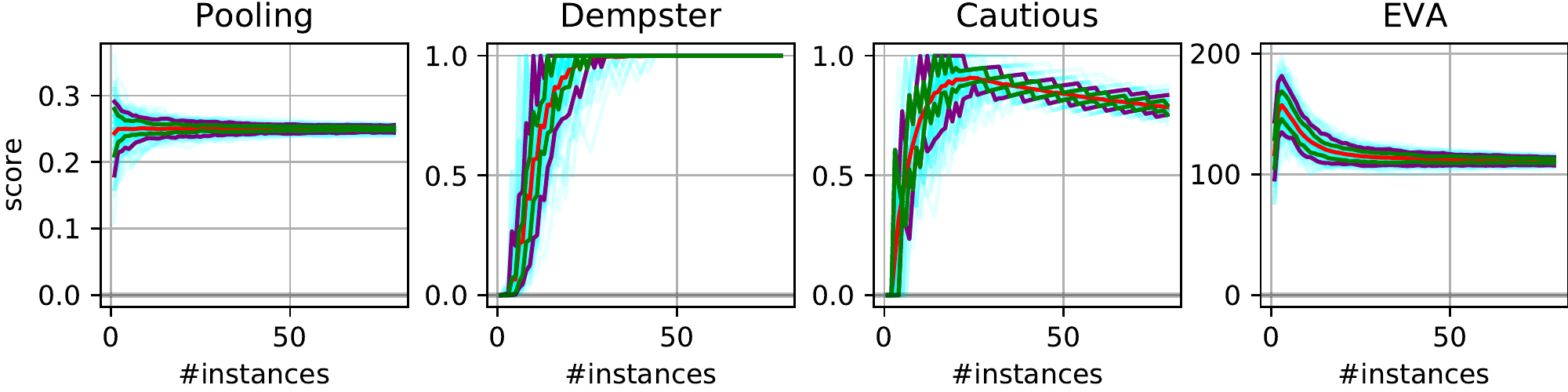} 
\caption{Evolution of the scores ($y$-axis) for each combination method based on simulating 
100 ensemble's final predictions (cyan lines) based each on 100 leafs sampled like in Figure~\ref{fig:prediction_methods}.
}
\label{fig:composite_methods}
\end{figure}

In contrast to the aggregation of scores, the integrated combination of leafs skips the intermediate step of score generation and directly combines the statistics of the leafs $\predaggfkt: \leafs^\nrtrees \rightarrow \mathbb{R}$ to form a prediction for the ensemble. Through this approach the exact statistics of the leafs can be considered for the aggregation which would otherwise be inaccessible using scores. 
In contrast to Figure~\ref{fig:prediction_methods}, Figure~\ref{fig:composite_methods} depicts the outputs of 100 simulated ensembles.

\subsubsection{Pooling.} 
The basic idea of pooling is to reduce the influence of leafs with a low number of instances which otherwise would obtain a high score either for the positive or the negative class. Therefore, this approach first sums up the leaf statistics and then computes the probability which can be defined as:
\begin{equation*}
    \fpool(\fvec) = \frac{\sum_{i=1}^{\nrtrees} \lpos_i}{\sum_{i=1}^{\nrtrees} \left( \lpos_i + \lneg_i \right)} -0.5
\end{equation*}

With respect to Figure~\ref{fig:composite_methods}, we can observe that pooling behaves similarly to the probability approach. The reduced variability is mainly caused due to the simulation of more leafs.

\subsubsection{Dempster.} 
Based on the \emph{Dempster-Shafer} framework, leaf aggregation can also be performed using the theory of belief functions~\citep{shafer1976mathematical}. 
For our purpose the states of belief can be defined as $\Omega=\{\plus,\minus\}$, where $\plus$ represents the belief for the positive and $\minus$ represents the belief for the negative class. A particular mass function $m: 2^\Omega \rightarrow [0,1]$ assigns a mass to every subset of $\Omega$ such that $\sum\nolimits_{A\subseteq\Omega}m(A)=1$. Based on the plausibility (c.f.\ Section~\ref{sec:plausibility}), we define the mass function for a particular leaf $\leaf$ as follows:
\begin{align*}
m_\leaf(\emptyset) &= 0 & m_\leaf(\{\plus\}) &= s_\plus(\leaf)\\\
m_\leaf(\{\minus\}) &= s_\minus(\leaf) & m_\leaf(\{\plus,\minus\}) &= u_e(\leaf) + u_a(\leaf)\text{.}
\end{align*}
In order to combine the predictions of multiple leafs, the mass functions can be aggregated using the (unnormalized) Dempster's rule of combination:
\begin{align*}
(m_{\leaf_1} \circled{$\cap$} m_{\leaf_2})(A)=\sum\limits_{B\cap C=A}m_{\leaf_1}(B)m_{\leaf_2}(C) && \forall A\subseteq\Omega
\end{align*}
Hence, for the ensemble we can define the following belief function:
\begin{align*}
    m_{dempster}(A)= (m_{\leaf_1} \circled{$\cap$} (m_{\leaf_2}  \circled{$\cap$} ( \ldots \circled{$\cap$} m_{\leaf_K}))) (A)
\end{align*}
which can be used to form a score for the positive class as follows:
\begin{align*}
    \fdem(\fvec) = m_{dempster}(\{\plus\}) - m_{dempster}(\{\minus\})
\end{align*}



\subsubsection{Cautious.} A particular drawback of Dempster's rule of combination is that it requires independence among the combined mass functions which is usually not true for classifiers in an ensemble which have been trained over the same data. The independence assumption can be omitted by using the cautious rule of combinations~\citep{denoeux2006cautious}. The idea behind the cautious rule is based on the \emph{Least Commitment Principle}~\citep{smets1993belief}  which states that, when considering two belief function, the least committed\com{ "informative"} one should be preferred. Following this principle, the cautious rule $\circled{$\land$}$ can be used instead of Dempster's rule of combination $\circled{$\cap$}$. For more information we refer to~\citet{denoeux2006cautious}.
Hence, for the ensemble we can define the following belief function
\begin{align*}
    m_{cautious}(A)= (m_{\leaf_1} \circled{$\land$} (m_{\leaf_2}  \circled{$\land$} ( \ldots \circled{$\land$} m_{\leaf_K}))) (A)
\end{align*}
which can be transformed to a score for the positive class as follows:
\begin{align*}
    \fcau(\fvec) = m_{cautious}(\{\plus\}) - m_{cautious}(\{\minus\})
\end{align*}

We use a small value ($10^{-5}$) as the minimum value for certain steps during the calculation to deal with numerical instabilities for both Dempster and Cautious. 

In order to provide a more intuitive explanation, consider the following three estimates to combine: $s_\plus(\leaf_1)=s_\plus(\leaf_2)=0.4$ and $s_\minus(\leaf_3)=0.4$. Dempster would consider that there are two estimates in favor of the positive class and hence predict it, whereas Cautious ignores several preferences for the same class, since they could result from dependent sources, and would just take the stronger one. Hence, 
Cautious would produce a tie (but high weights for $\emptyset$ and $\{\plus,\minus\}$).
This behaviour can also be observed in Figure~\ref{fig:composite_methods} where the plot for Cautious resulted very similar to that of the maximum operator (not shown  due to space restrictions). 
For Dempster, in contrast, the majority of leafs confirming the positive class very quickly push the prediction towards the extreme. 
Consider for that  the following scenario, which also visualizes the behaviour under conflict. For $s_\plus(\leaf_1)=s_\plus(\leaf_2)=0.8$ and $s_\minus(\leaf_3)=0.98$ Dempster's rule would prefer the negative over the positive class, but already give a weight of 0.94 to neither $\plus$ nor $\minus$, whereas Cautious would prefer the negative class with a score of 0.20 and 0.78 for $\emptyset$.



\subsubsection{Evidence accumulation.} 
In the presence of uncertainty, it may make sense to reach a decision not by combining individual recommendations themselves, but by accumulating the {\em evidence} underlying each recommendation. This motivates a new combining rule, here called EVA (EVidence Accumulation).

To get the intuition behind the idea, consider a police inspector investigating a crime. It is perfectly possible that no clue by itself is convincing enough to arrest a particular person, but jointly, they do. This effect can never be achieved by expressing the evidence as probabilities and then averaging.  E.g., assume there are 5 suspects, A, B, C, D, E; information from Witness $\leaf_1$ rules out D and E, and information from Witness $\leaf_2$ rules out A and B.  Logical deduction leaves C as the only possible solution. Now assume $\leaf_1$ and $\leaf_2$ express their knowledge as probabilities: $\leaf_1$ returns  (0.33, 0.33, 0.33, 0, 0) (assigning equal probabilities to A, B and C and ruling out D and E), and $\leaf_2$ returns (0, 0, 0.33,0.33, 0.33). Averaging these probabilities gives (0.17, 0.17, 0.33, 0.17, 0.17), when logic tells us it should be (0, 0, 1, 0, 0). Note that 1 is out of the range of probabilities observed for C: no weighted average can yield 1.  Somehow the evidence from different sources needs to be {\em accumulated}.

In probabilistic terms, the question boils down to: How can we express $P(y \given \bigwedge \leaf_i)$ in terms of $P(y \given \leaf_i)$?  Translated to our example, how does the probability of guilt, given the joint evidence, relate to the probability of guilt given each individual piece?
There is not one way to correctly compute this: assumptions need to be made about the independence of these sources.  Under the assumption of class-conditional independence (as made by, e.g., Naive Bayes), we can easily derive a simple and interpretable formula:



\begin{align*}
P(y \given \bigwedge_i \leaf_i) \sim P(\bigwedge \leaf_i \given y) \cdot P(y)   =  P(y) \prod_i P(\leaf_i \given y)  \sim P(y) \prod_i \frac{P(y \given \leaf_i)}{P(y)} 
\end{align*}


Thus, under class-conditional independence, evidence from different sources should be accumulated by multiplying the prior $P(Y)$ with a factor $P(y \given \leaf_i) / P(y)$ for each source $\leaf_i$. That is, if a new piece of information, on its own, would make $y$ twice as likely, it also does so when combined with other evidence.
In our example, starting from $(0.2, 0.2, 0.2, 0.2, 0.2)$, $\leaf_1$ multiplies the probabilities of A, B, C by 0.33/0.2 and those of D, E by 0, and $\leaf_2$ 
lifts C, D, E by 0.33/0.2 and A, B by 0; this gives $(0, 0, 0.54, 0, 0)$ which after normalization becomes $(0, 0, 1, 0, 0)$.

Coming back to the binary classification setting, we compute
\begin{equation*}
\feva(\fvec)= P(\plus) \prod_i \frac{P(\plus \given \leaf_i)}{P(\plus)} - P(\minus) \prod_i \frac{P(\minus \given \leaf_i)}{P(\minus)}
\end{equation*}
where $P(\plus)$ and $P(\minus)$ is estimated by the class distribution on the training data and $P(y \given \leaf_i)$ essentially comes down to $\pprob(\leaf_i)$.
We apply similar tricks as in Naive Bayes for ensuring numerical stability such as a slight Laplace correction of 0.1 in the computation of $P(y \given \leaf_i)$.

With respect to Figure~\ref{fig:composite_methods}, we can observe an even stronger decrease of the absolute score with increasing leaf size than for Cautious. 
This indicates that the output of EVA can be highly influenced by smaller leafs which are more likely to carry more evidence than larger leafs, in the sense that their estimates are more committed.

\section{Evaluation \com{4 pages. Status: almost finished}}
\label{sec:evaluation}

A key aspect of our experimental evaluation is to compare our proposed methods for combining predictions with respect to the conventional approaches of voting and averaging probabilities. In order to put the combination strategies to the test, our experiments are based on random decision trees which, in contrast to other tree ensembles, guarantee a high diversity of estimates in the ensemble.

\subsection{Experimental Setup}
\label{sec:setup}
\label{sec:experiment}

\begin{table}[t]
    \centering
    \caption{Binary classification datasets and statistics. 
    }
    \label{tab:datasets}
    \resizebox{1.0\textwidth}{!}{%
    \begin{tabular}{| c | c | c | c || c | c | c | c |}
         \hline
        name & \#instances & \#features & class ratio & name & \#instances & \#features & class ratio\\ 
        \hline
        scene & 2407 & 299 &         0.18 & sonar & 208 & 60 &           0.47 \\ 
        webdata & 36974 & 123 &      0.24 & mushroom & 8124 & 22 &       0.48 \\
        transfusion & 748 & 4 &      0.24 & vehicle & 98528 & 100 &      0.50 \\
        biodeg & 1055 & 41 &         0.34 & phishing & 11055 & 30 &      0.56 \\
        telescope & 19020 & 10 &     0.35 & breast-cancer & 569 & 30 &   0.63 \\
        diabetes & 768 & 8 &         0.35 & ionosphere & 351 & 34 &      0.64 \\
        voting        & 435 & 16 &   0.39 & tic-tac-toe & 958 & 9 &      0.65 \\
        spambase & 4601 & 57 &       0.39 & particle & 130064 & 50 &     0.72 \\
        electricity & 45312 & 8 &    0.42 & skin & 245057 & 3 &          0.79 \\
        banknote      & 1372 & 4 &   0.44 & climate & 540 & 20 &         0.91 \\
        airlines      & 539383 & 7 & 0.45 & & & & \\ 
        \hline
    \end{tabular}
    }
\end{table}


Our evaluation is based on $21$ datasets for binary classification\footnote{Downloaded from the UCI Machine Learning Repository \url{http://archive.ics.uci.edu} and OpenML \url{https://www.openml.org/}.}, 
shown in Table~\ref{tab:datasets}, which we have chosen in order to obtain diversity w.r.t. the size of dataset, the number features and the balance of the class distribution. To consider a variety of scenarios for combining unstable predictions, we have performed all of our evaluations with respect to an ensemble of $100$ trees and minimum leaf sizes 
1, 2, 3, 4, 8 or 32.
Note that a single RDT ensemble for each leaf size configuration is enough in order to obtain the raw counts $\leaf_i$ and hence produce the final predictions for all methods. 
We used 5 times two fold cross validation in order to decrease the dependence of randomness on the comparability of the results.%
\footnote{Our code is publicly available at \url{https://github.com/olfub/RDT-Uncertainty}.} 

For measuring the performance, we have computed the area under receiver operating characteristic curve (AUC) and the accuracy. For a better comparison we have computed average ranks across all datasets. As an additional reference points to our comparison, we added the results of a single decision tree and a random forest ensemble. Note that though imposing an equal tree structure, these trees were trained with the objective of obtaining a high purity of class distributions in their leaves. Therefore, we did not expect to reach the performance w.r.t. classification but it was intended to show to what degree advanced combination strategies are able to close the gap between using  tree models resulting merely from the data distribution and tree models specifically optimized for a specific task. 


\begin{figure}[t!]
    \centering
    \resizebox{\textwidth}{!}{%
        \includegraphics[height=5cm]{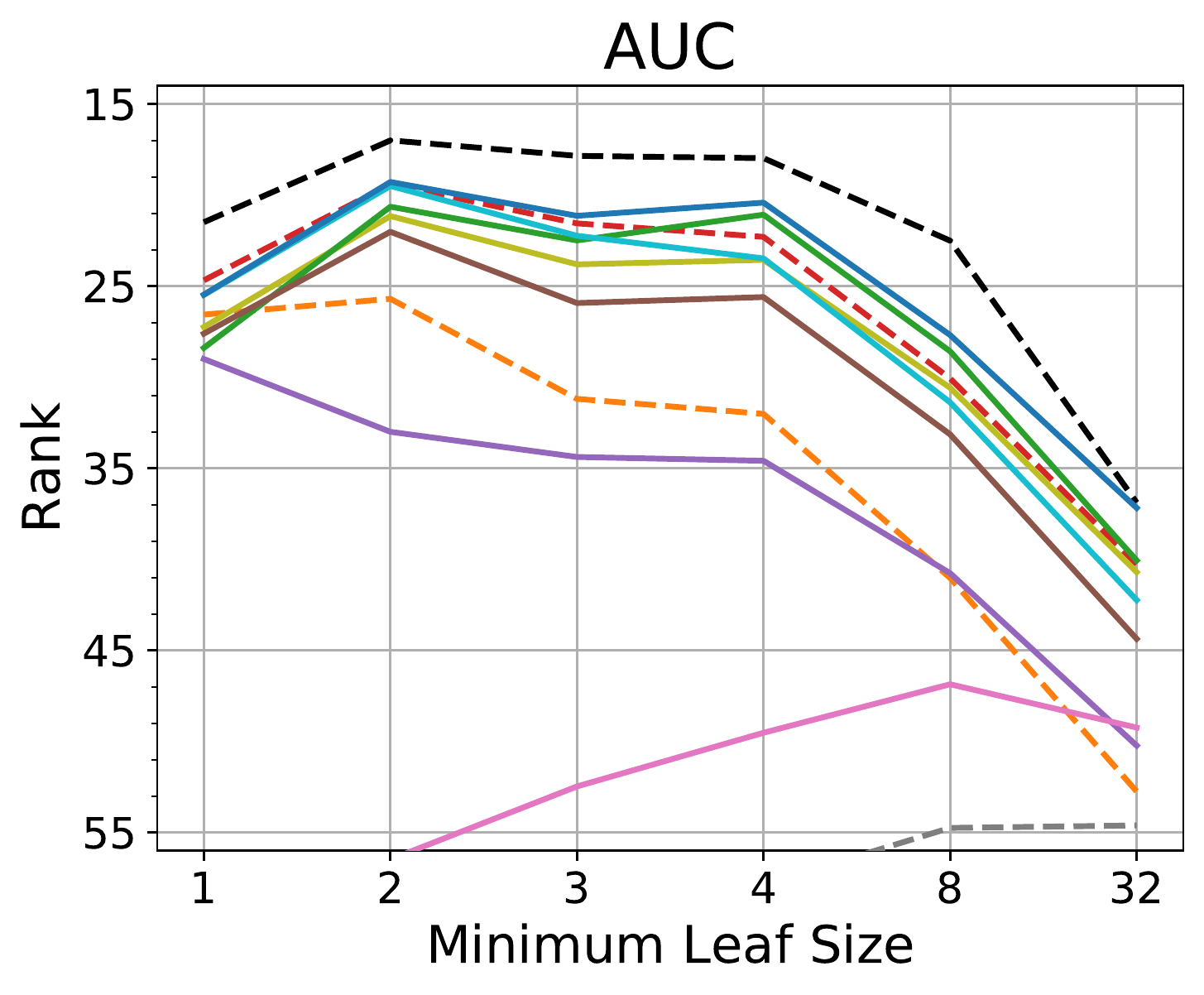}
        \includegraphics[height=5cm]{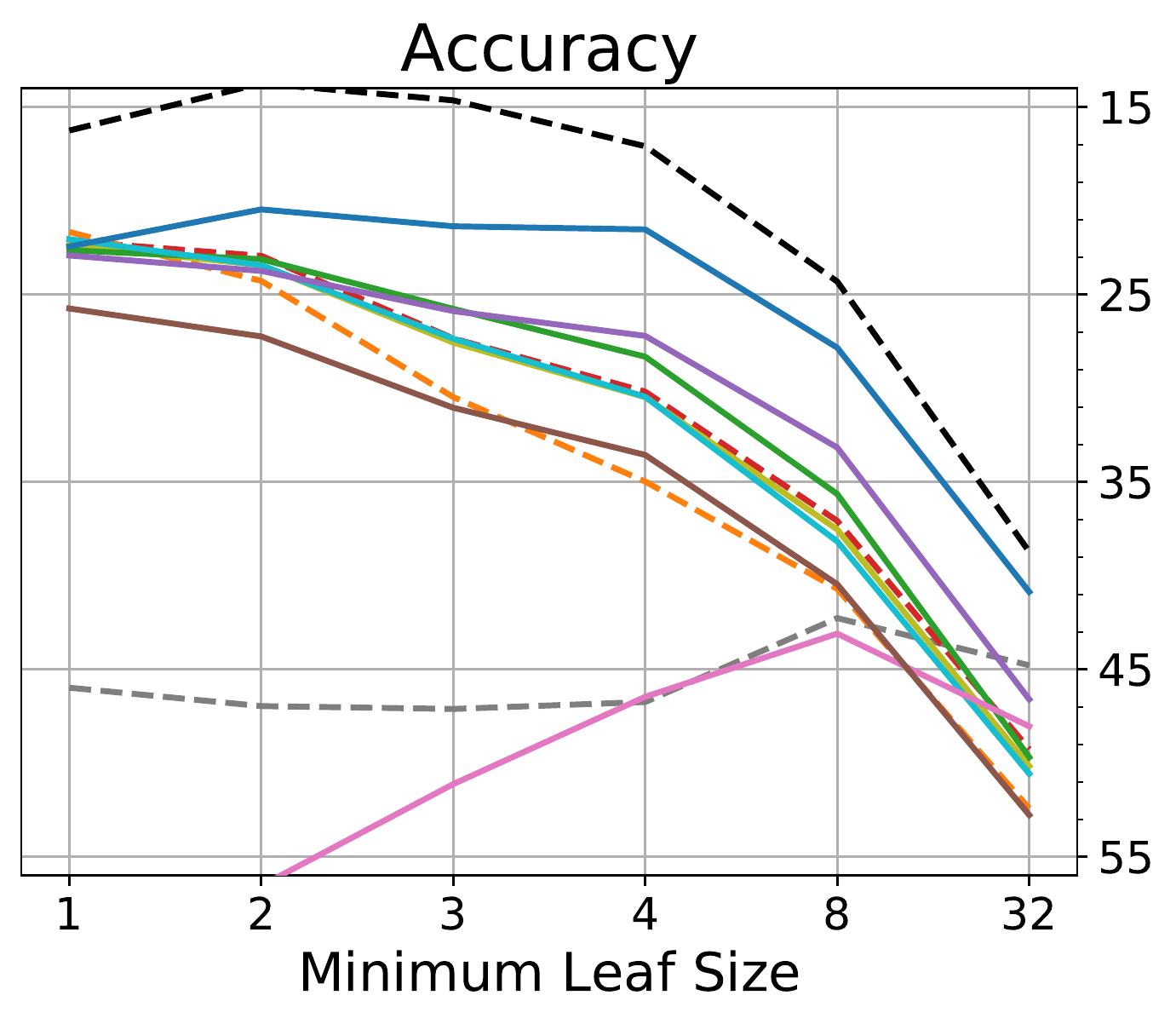}
    }\\
    \includegraphics[width=\linewidth]{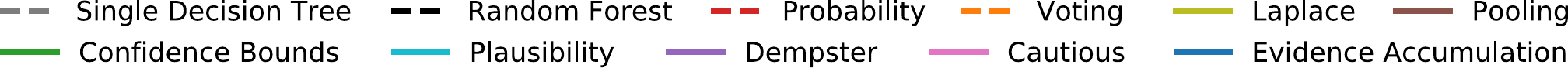}\\
    \caption{
    Comparison of the average ranks with respect to the 11 methods and the 6 leaf configurations (worst rank is $6\cdot 11=66$). Left: AUC. Right: Accuracy.
    }
    \label{fig:eval1}
\end{figure}

\subsection{Results}
\label{sec:analysis}

\begin{figure}[th]
    \centering
    \resizebox{\textwidth}{!}{%
        \includegraphics[height=4cm]{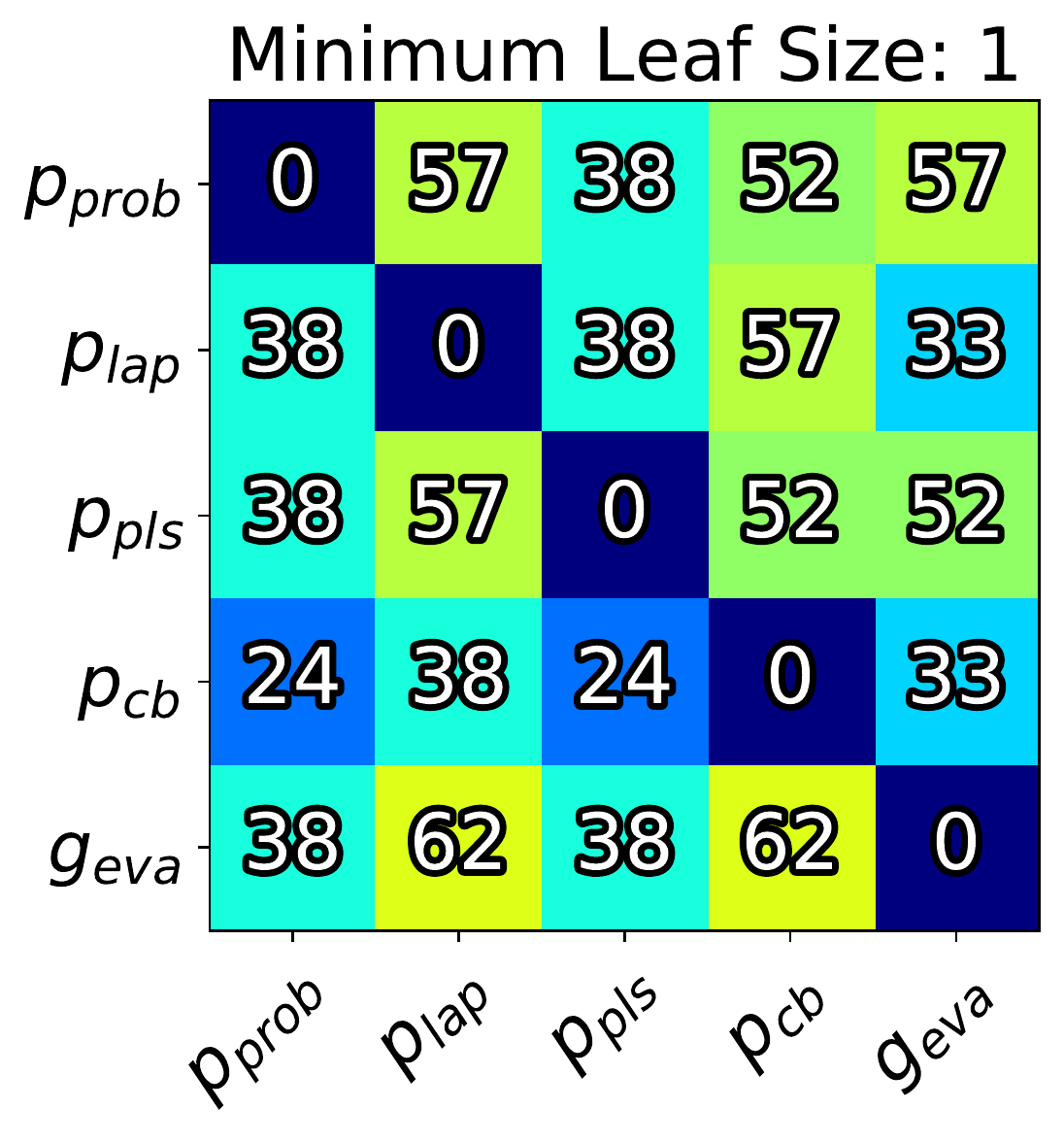}
        \includegraphics[height=4cm]{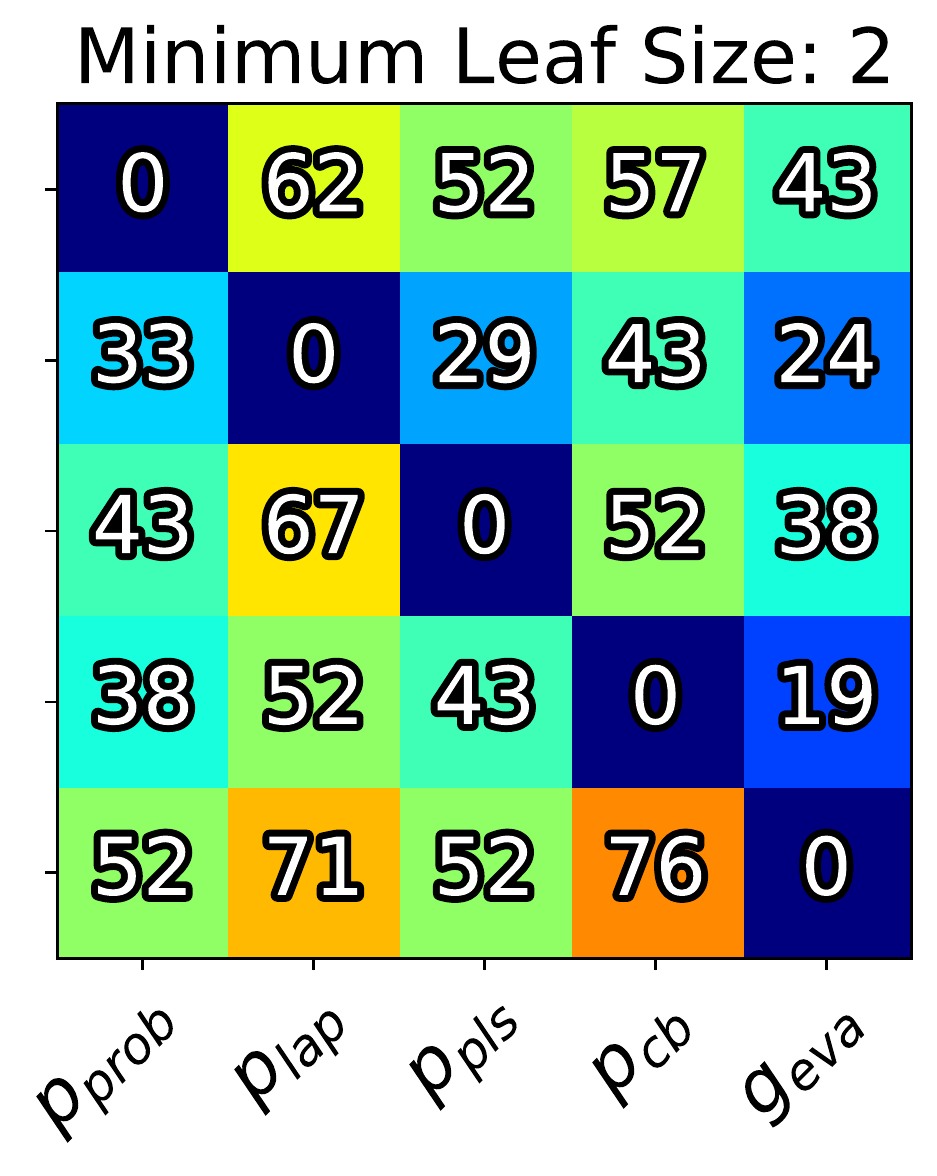}
        \includegraphics[height=4cm]{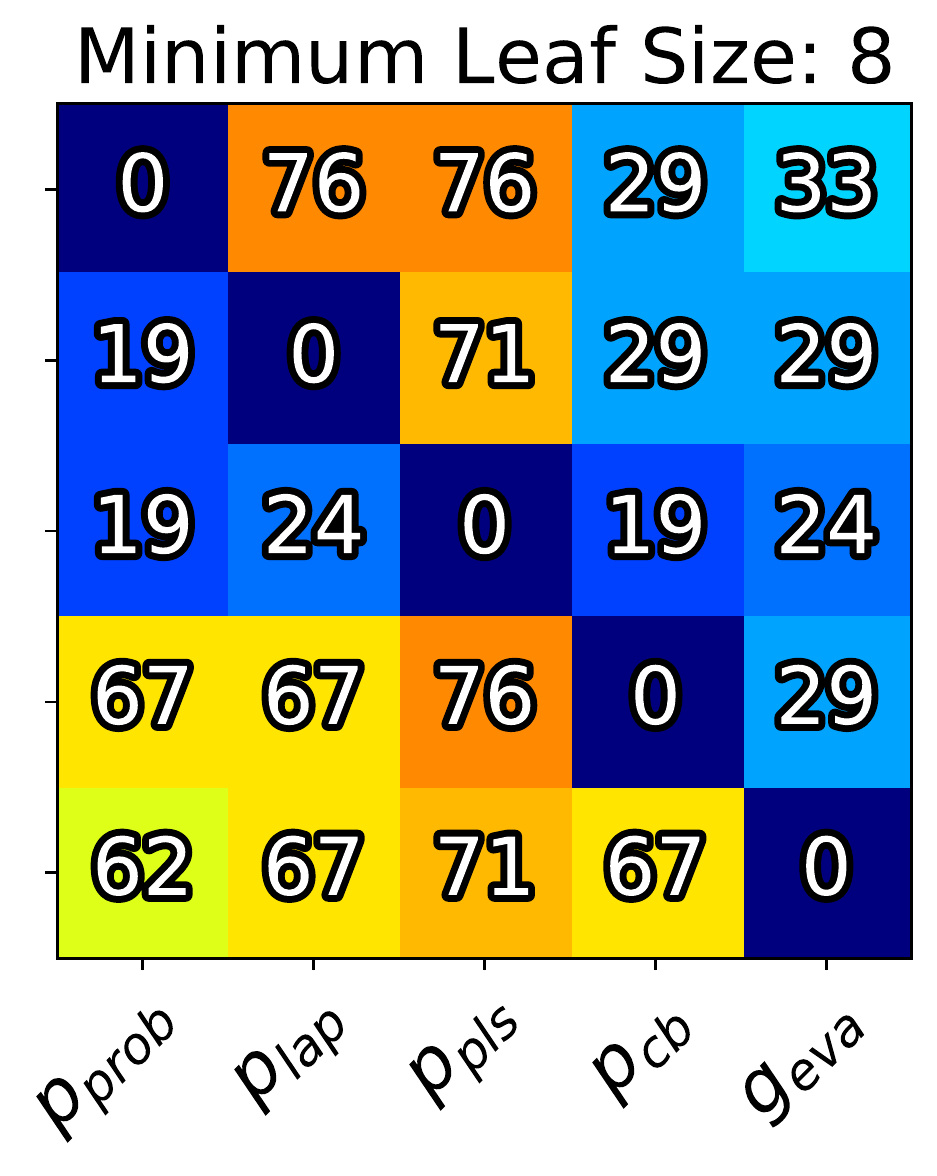}
        \includegraphics[height=4cm]{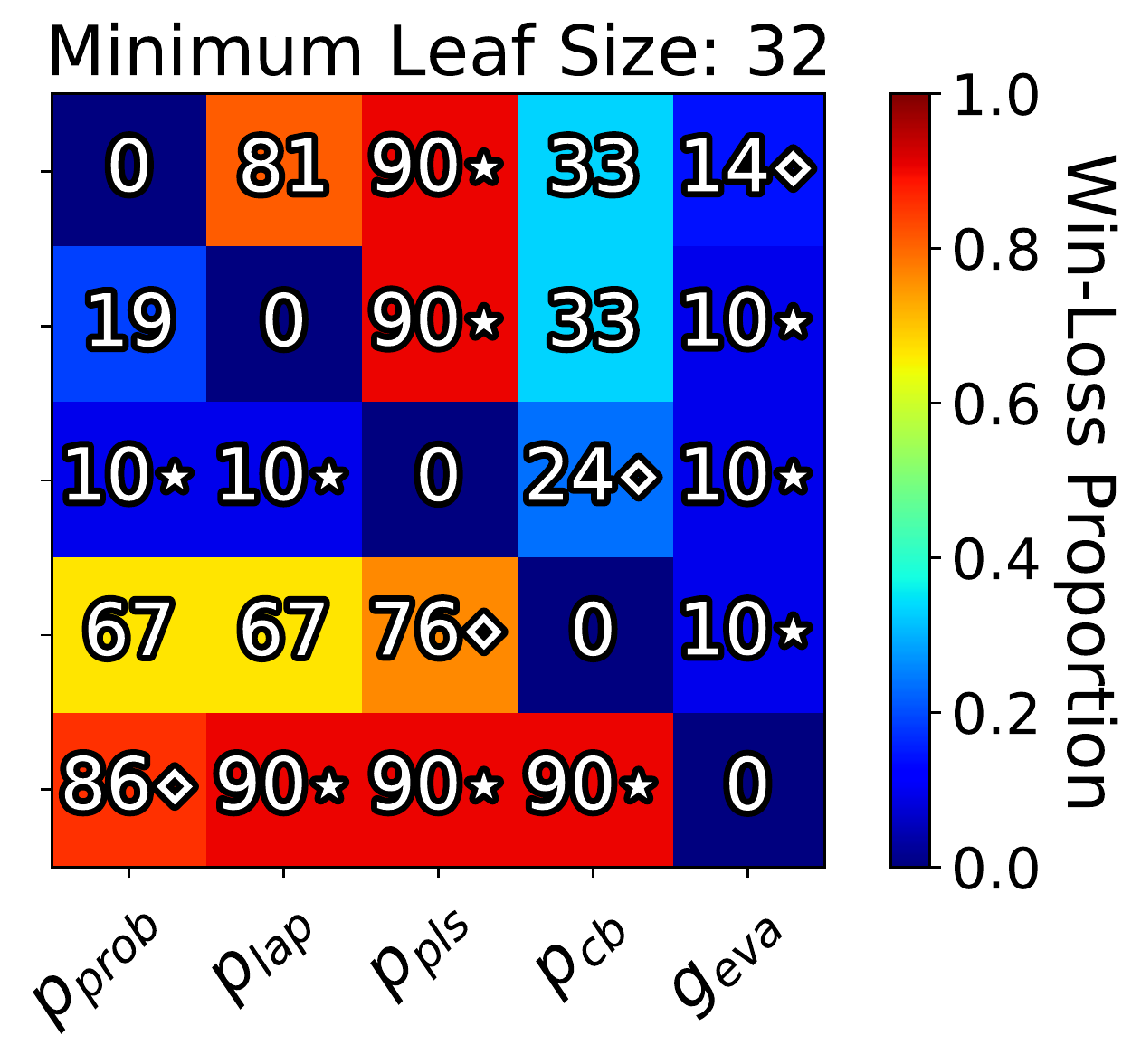}
    }\\
    \caption{
    Heatmap of pairwise comparisons. Each row and each column belongs to a method in the order $\pprob$, $\plap$, $\ppls$, $\pcb$, and $\feva$ and the number and color indicates how often the method in the row had a better AUC score than the method in the column. $\bullet$, $\diamond$ and $\star$  indicate a  significant difference according to the Bonferroni corrected Wilcoxon signed-rank test with $\alpha=0.05$, 0.01 and 0.001, respectively.
    }
    \label{fig:best5_heatmap}
\end{figure}

Figure~\ref{fig:eval1} shows a comparison between all methods with respect to different tree structures.
The first observation is that  most of the methods perform around the baseline of taking the average probability.
Exceptions are evidence accumulation, which is on top of the remaining approaches especially in terms of accuracy, and the approaches based on the theory of belief functions, which especially exhibit problems in producing rankings (AUC). 
Nonetheless, Dempster rule of combination has an advantage over the other methods for middle to large sized leafs in terms of accuracy. 
Figure~\ref{fig:best5_heatmap} shows again how close the methods are together (Dempster, Cautious, Pooling and Voting were left out).
We can see that $\plap$ always is worse than $\pprob$, regardless of the leaf size.
While $\ppls$ is not much worse than the best method ($\pprob$) on small leafs, it falls behind with increasing leaf size.
Without $\feva$, $\pcb$ would be the best method on medium sized or larger leafs but $\feva$ has even better results there.
Overall, $\pprob$ most often has the best results on very small leafs, but for not much larger leafs, $\pprob$ starts falling behind $\feva$. 

\begin{figure}[t]
    \centering
    \includegraphics[width=0.49\linewidth]{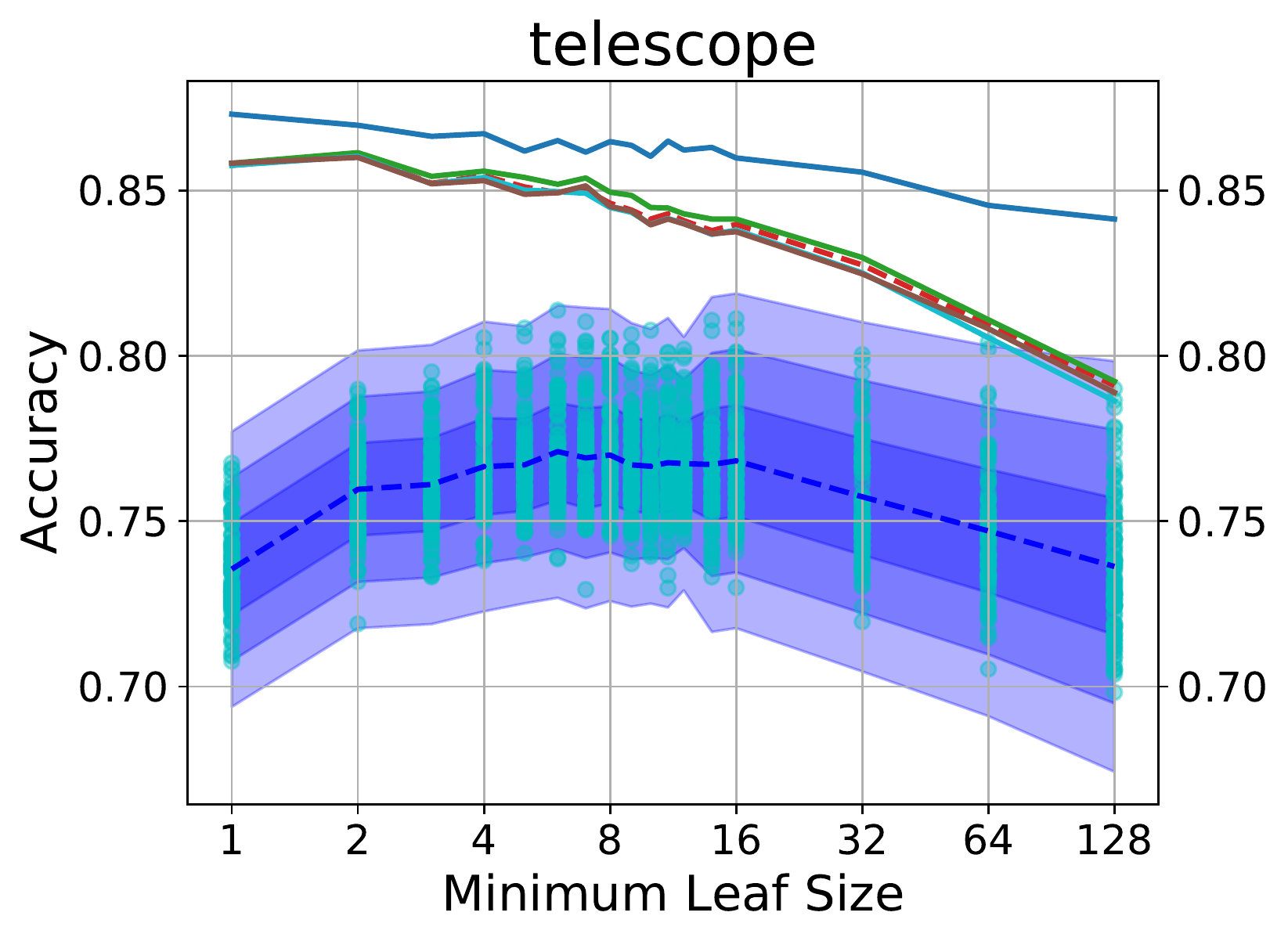}
    \includegraphics[width=0.49\linewidth]{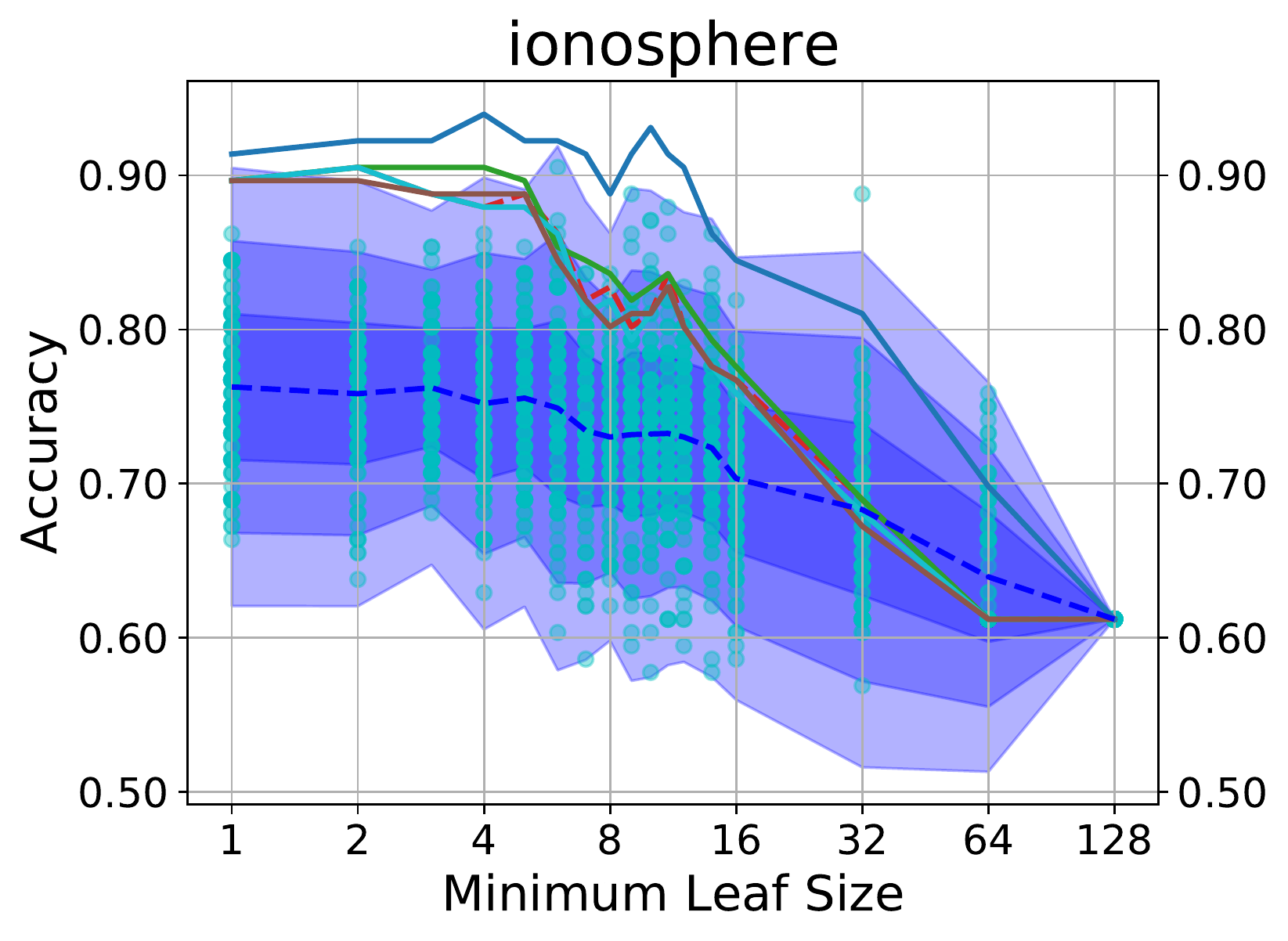}
    \includegraphics[width=0.49\linewidth]{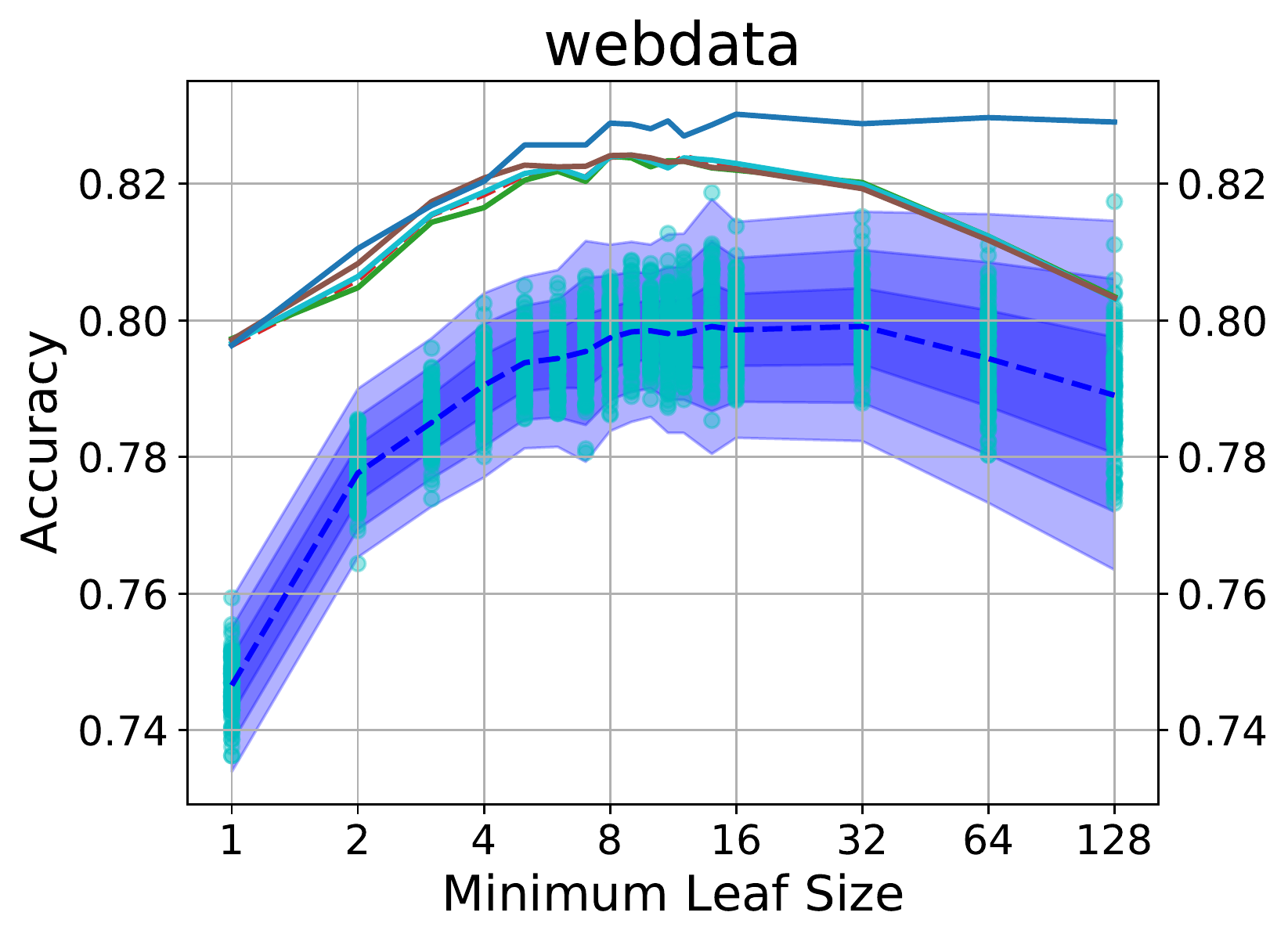}
    \includegraphics[width=0.49\linewidth]{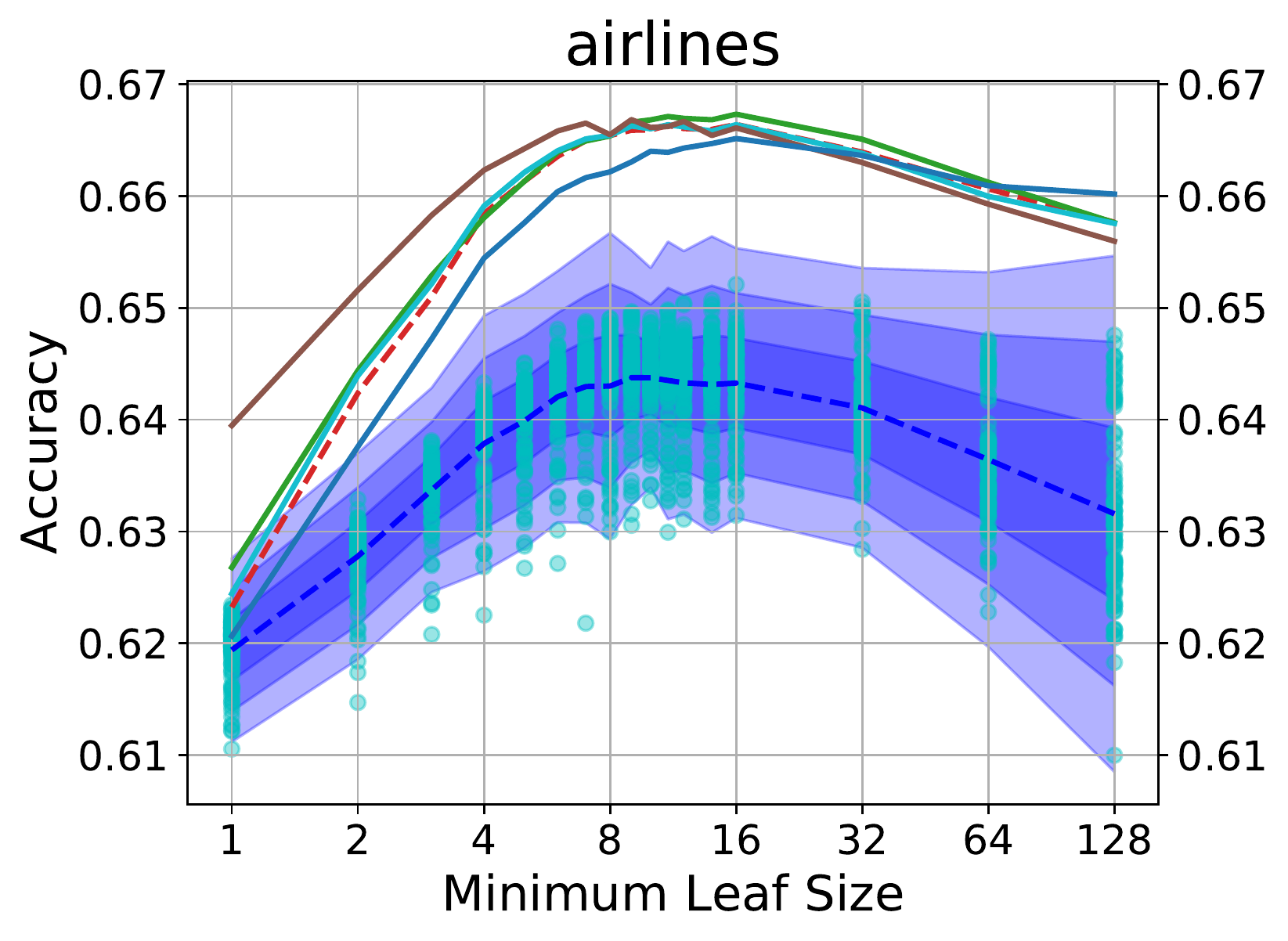}
    \includegraphics[width=\linewidth]{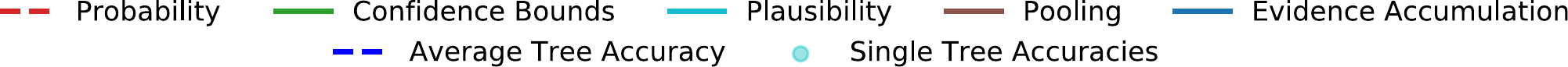}
    \caption{Accuracy of selection of five combination strategies and distribution with mean (dashed blue line), 1, 2 and 3 standard deviations (blue shaded areas) over accuracies of the underlying individual trees in the ensemble (turquoise dots).}
    \label{fig:datasets}
\end{figure}

Figure~\ref{fig:datasets} provides further insights on four selected datasets (single random 50\% train/test split). It becomes apparent that the advantage of EVA is also often substantial in absolute terms. 
On the other hand, we can also observe cases where EVA falls behind the other approaches (\emph{airlines}).
Interestingly,  the accuracy of EVA on \emph{webdata} increases with increasing leaf size long after the other methods reach their peak, contrary to the general trend.

We can further observe that, as expected, the performances of the proposed methods lay between those of random forests and the single decision tree. 
However, Figure~\ref{fig:class_ratio} reveals that an important factor is the class ratio of the classification task. 
While random forest clearly outperform all other methods for highly imbalanced tasks, the advance is negligible for the balanced problems. 
This indicates a general problem of RDT with imbalanced data, since it gets less likely on such data to obtain leafs 
with (high) counts for the minority class.

\subsection{Discussion\todo{status: almost finished}}
In the following, we 
discuss the results of each individual method in more detail. 
\subsubsection{Probability, Laplace, Voting and Pooling.}

Computing the expectation over $\pprob$ straightforwardly\com{I wanted to show that we we know what this actually is doing, see also the next outcommented line}
showed to be very effective compared to most of the alternative method ideas.
Especially when small leafs are involved this behaviour was not expected since we believed that small leafs with overly certain but likely wrong estimates (probabilities 0 and 1) would out-weight the larger leafs which provide more evidence for their smoother estimates. 
Instead, the results indicate that a large enough ensemble (we used 100 trees) makes up for what we considered too optimistic $\pprob$ leaf predictions \citep{prob_random_forest}\todo{that was the ref, right?}.
In fact, applying Laplace correction towards uniform distribution
was hindering especially for small leaf sizes in our experiments. 
\com{falls class ratio grafik: however, the correction showed a slight advantage for imbalanced data}
\com{Perhaps include something like: From a probabilistic point of view this is reasonable, since averaging the $\pprob$ corresponds to computing the expected value 
of different realizations of $P(y \ given x)$.}
Voting is able to catch-up with the other methods only for the configuration of  trees with very small leafs, where, indeed, the scores to be combined are identical to those of $\pprob$ except for leafs with size greater one.
Interestingly, Pooling improves over the other methods on a few datasets as it can be seen for \emph{airlines} in Figure~\ref{fig:datasets}. 
Nevertheless, the results on most of the datasets indicate that the consideration of the leaf size by this method might be too extreme in most cases, leading generally to unstable results. 

\begin{figure}[t]
    \centering
    \resizebox{\textwidth}{!}{%
        \includegraphics[height=5cm]{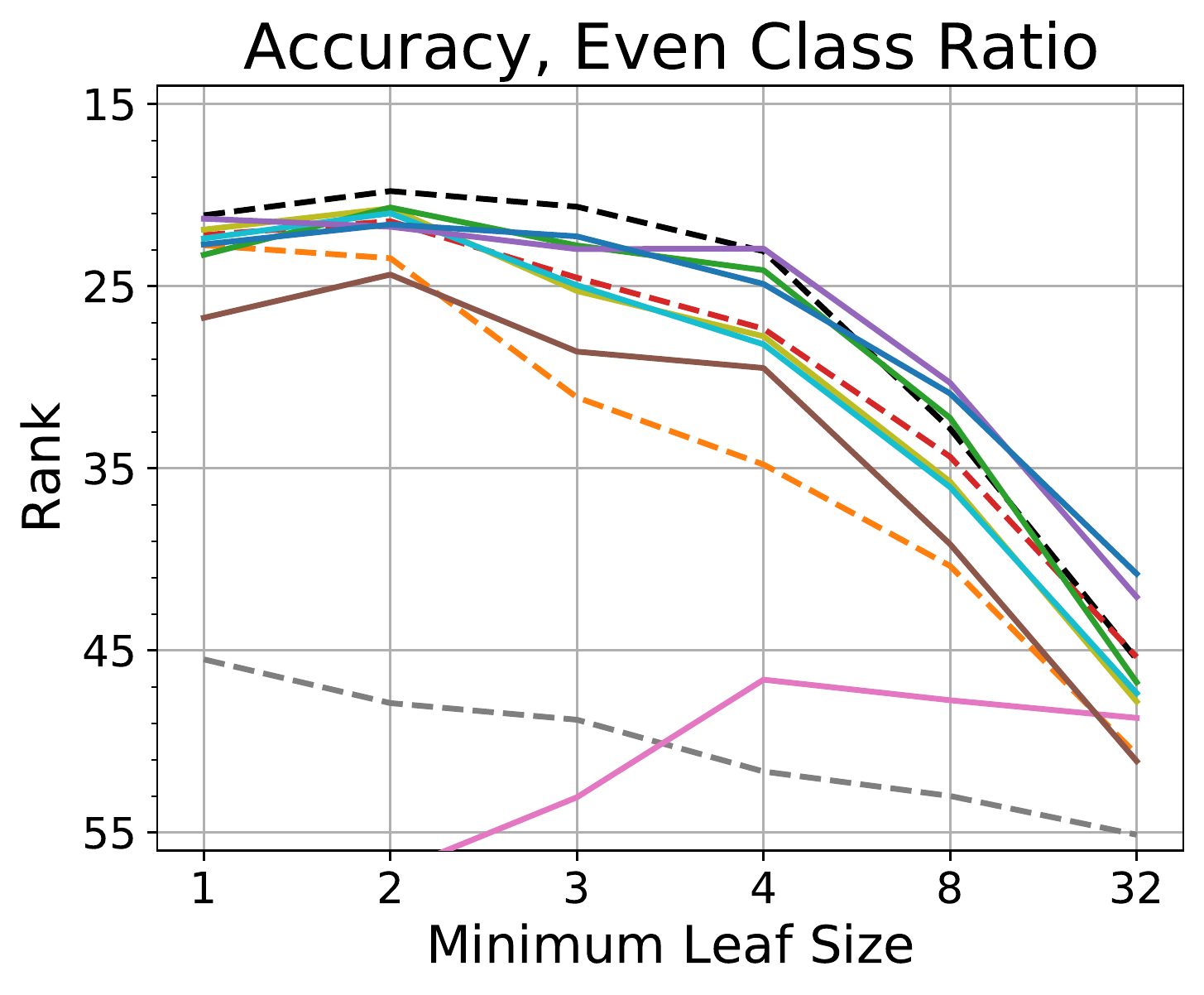}
        \includegraphics[height=5cm]{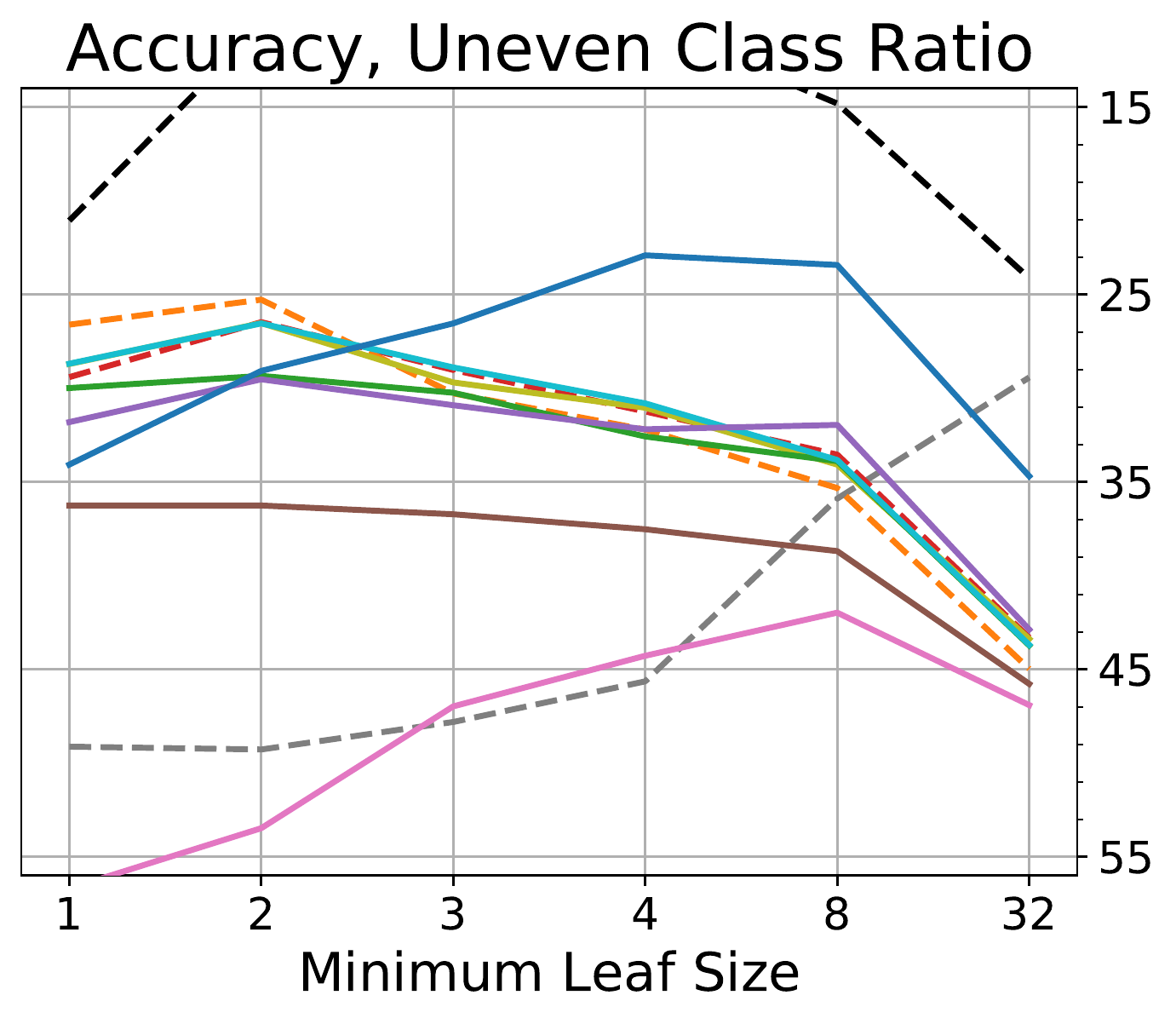}
    }\\
    \includegraphics[width=\linewidth]{figures/evaluation/paper_default_legend.pdf}\\
    \caption{Accuracy ranks on dataset with relatively even class ratio (0.4 to 0.6) on the left and relatively uneven class ratio (less than 0.3 or greater 0.7) on the right.
}
    \label{fig:class_ratio}
\end{figure}

\subsubsection{Plausability, Dempster and Cautious.}
We expected that a precise computation and differentiation of plausibility and uncertainties would be beneficial in particular for small leaf sizes. However, we were not able to observe a systematic advantage of $\ppls$ over the $\pprob$ baseline in our experiments. 
Nonetheless, the usage of the uncertainty estimates in Dempster's rule of combinations shows the potential of having access to such  scores.
When the data was balanced, Dempster classified as good or better than EVA or even the random forest.
The gap w.r.t. AUC demonstrates that the proposed translation of the components of the mass function into one score is not yet optimal.\com{we could also mention, that we may use \{\} and +- scores for something, e.g. for abstaining}
Moreover, the results suggests that the lack of independence  is not problematic when combining predictions in RDT, but on the contrary, the way the Cautious rule addressed it caused serious problems.

\subsubsection{Confidence Bounds.}
Among the methods with similar performance to $\pprob$, the approach using confidence bounds exhibits the greatest (but still small) advantage over the baseline for trees with medium to large leaf sizes. 
For very small leafs, however, the chosen statistical modeling is obviously not ideal, at least for ranking. 
Also, it might be possible to achieve even better results with the same idea by using distributions or more elaborated strategies which can achieve a better fit. 

\subsubsection{Evidence Accumulation.}
Using a small value for Laplace smoothing, $\feva$ proved to be a very effective prediction strategy across all but the very smallest leafs.
This suggests that the assumptions underlying EVA are well-met by RDTs. The randomness of the leaves fits EVA's independence assumption.

The stronger consideration of the class prior probability than in other methods may also play a role on why EVA works better on unbalanced data, particularly for larger leaf sizes (Figure~\ref{fig:class_ratio}). 
Remind that EVA relates the estimates for each source to the prior probability and judges the evidence according to how much the leaf distribution deviates from it.
Leafs for the minority class might hence have a higher impact on the final prediction than for other methods, where such leafs are more likely to be out-ruled by the average operation and the leafs following the class prior. 


\section{Conclusions \com{max. 2 pages with references. Status: still initial draft}}
\label{sec:conclusion}

In this work, we proposed methods to combine predictions under uncertainty in an ensemble of decision trees. \emph{Uncertainty} here refers not only to how uncertain a prediction is, but also to uncertainty about this uncertainty estimate. Our methods include Laplace smoothing, distinguishing between aleatoric and epistemic uncertainty, making use of the dispersion of probability distributions,  combining uncertainties under the theory of belief functions, and accumulating evidence. 
Random decision trees ensured a high diversity of the predictions to combine and a controlled environment for our experimental evaluation.
We found that including the support for the available evidence in the combination improves performance over averaging probabilities only in some specific cases. 
However, we could observe a consistent advantage for the proposed approach of evidence accumulation.

This result suggests that  the strength of the evidence might be a better factor than the support for the evidence when combining unstable and diverse predictions from an ensemble. 
However, the principle of evidence accumulation 
revealed additional aspects,
like the rigorous integration of the prior probability or the multiplicative accumulation of evidence, which might also be relevant for the observed improvement. 
Moreover, Dempster's rule showed that combining support (through the plausibilities) and commitment of evidence produces good results when certain conditions, such as even class ratio, are met.
We plan to investigate these aspects in future work, e.g. by integrating a prior into Dempster's rule or aspects of reliable classification into EVA. 
Further research questions include which insights can be carried over to random forest ensembles, which are more homogeneous in their predictions, or 
if the presented strategies  have a special  advantage 
under certain conditions, such as  small ensembles.



%
%
%
%
\bibliographystyle{splncs04nat}
\footnotesize
\bibliography{bib}


\end{document}